%% file: main.tex
\documentclass[10pt,twocolumn,letterpaper]{article}

\usepackage{cvpr}              % To produce the CAMERA-READY version
\usepackage[dvipsnames, table]{xcolor} % add

\usepackage[]{color-edits}

\input{preamble}

\definecolor{cvprblue}{rgb}{0.21,0.49,0.74}
\usepackage[pagebackref,breaklinks,colorlinks,citecolor=cvprblue]{hyperref}

\usepackage{multirow}
\usepackage{blindtext}
\usepackage{tcolorbox}
\usepackage{graphicx} 
\usepackage{pifont}% http://ctan.org/pkg/pifont
\newcommand{\xmark}{\ding{55}}%
\def\paperID{1512} % *** Enter the Paper ID here
\def\confName{CVPR}
\def\confYear{2024}

\title{VLP: Vision Language Planning for Autonomous Driving}
\author{Chenbin Pan$^{1,2}$ \ \  Burhaneddin Yaman \ Tommaso Nesti$^{2}$ \ Abhirup Mallik$^{2}$\\ \ Alessandro G Allievi$^{2}$ \ Senem Velipasalar$^{1}$ \ Liu Ren$^{2}$\\
$^{1}$Syracuse University \ \ \ \\ $^{2}$Bosch Research North America \& Bosch Center for Artificial Intelligence (BCAI)\\
{\tt\small \{cpan14, svelipas\}@syr.edu}, \ \ \ \\  {\tt\small \{burhaneddin.yaman,tommaso.nesti,abhirup.mallik,alessandro.allievi,liu.ren\}@us.bosch.com}
}

\begin{document}
\maketitle

\input{sec/0_abstract}
\input{sec/1_intro_new}
\input{sec/2_related}
\input{sec/3_model}

\input{sec/4_exp}
\input{sec/5_conclusion}
\input{sec/X_suppl}

% \section{Methodology}

% Incorporates Large Language Models (LLMs) for contrastive learning during training and performs vision-only inference,

% \begin{itemize}
%     \item LLM with AD boost performance
%     \item Agent-wise learning within individual examples enhance the semantic representation of the feature map locally. Alleviate the batch-size requirement of contrastive learning.
%     \item LLM improve the generalization ability. (benefits compared with traditional data augmentation method: Contextual Embeddings, Language-Driven Generalization, Semantic Alignment, Efficient Feature Learning, Adaptability to Open-World Scenarios, Reduced Overfitting, no need for additional data annotation.)
% \end{itemize}
% \input{sec/0_abstract}    
% \input{sec/1_intro}
%\input{sec/2_formatting}
%\input{sec/3_finalcopy}
{
    \small
    \bibliographystyle{ieeenat_fullname}
    \bibliography{main}
}


\end{document}

%% file: preamble.tex
\usepackage[dvipsnames]{xcolor}

%% file: sec/0_abstract.tex
\begin{abstract}
Autonomous driving is a complex and challenging task that aims at safe motion planning through scene understanding and reasoning. While vision-only autonomous driving methods have recently achieved notable performance, through enhanced scene understanding, several key issues, including lack of reasoning, low generalization performance and long-tail scenarios, still need to be addressed. In this paper, we present VLP, a novel Vision-Language-Planning framework that exploits language models to bridge the gap between linguistic understanding and autonomous driving. VLP enhances autonomous driving systems by strengthening both the source memory foundation and the self-driving car's contextual understanding. VLP achieves state-of-the-art end-to-end planning performance on the challenging NuScenes dataset by achieving  35.9\% and 60.5\% reduction in terms of average L2 error and collision rates, respectively, compared to the previous best method. Moreover, VLP shows improved performance in challenging long-tail scenarios and strong generalization capabilities when faced with new urban environments.
\end{abstract}

%% file: sec/1_intro_new.tex
\section{Introduction}
\label{sec:1_intro}
Autonomous driving is a complex problem requiring scene understanding and reasoning to ensure safe motion planning. This sophisticated challenge can be broadly divided into three main tasks, namely perception, prediction and planning (also known as P3). Conventional methods adopt a modular approach by developing and optimizing each task in a disjoint manner without a holistic view, leading to compounding errors and safety concerns~\cite{STP3,Uniad,VAD}. End-to-end autonomous driving systems (ADS), unifying all P3 tasks, have garnered attention for their potential to enhance safe planning. Existing vision-based ADS typically follow a two-stage process:~bird's eye view (BEV) feature extraction and downstream tasks~\cite{Uniad, VAD}.~BEV feature extraction transforms multi-view camera data into a structured top-view representation embedding spatial information around the ego-car.~BEV features are further utilized as the information pool by the downstream P3 tasks. While effective, these vision-only methods struggle with out-of-domain generalization, such as maintaining performance in new cities and long-tail scenarios, hindering real-world deployment. 

The way humans process and interpret visual driving scenes naturally involve a coherent cognitive framework that maintains a consistent logical flow. This enables humans effortlessly make correct decisions even when faced with previously unseen data or scenarios, eliminating generalization issues in driving tasks. Then, this raises a pertinent question: \textit{How can ADS achieve human-like driving to navigate diverse and dynamic road environments?}

\begin{figure}[bt!]
  \centering
   \includegraphics[width=1.0\linewidth]{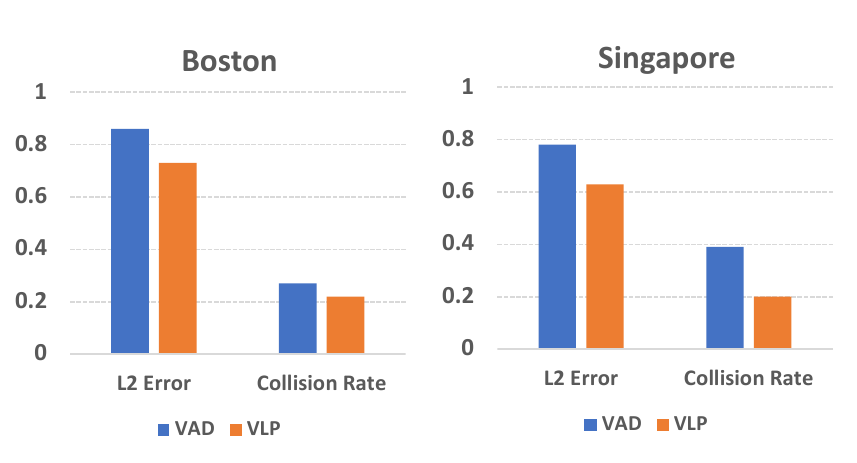}
   \vspace{-0.8cm}
   \caption{New-city generalization ability of ADS for planning is evaluated by training on Boston city and testing on Singapore, and vice versa. Our proposed VLP shows strong generalization ability by significantly outperforming state-of-the-art vision-only method, VAD\cite{VAD}, in terms of both L2 error and collision rate. }
   \label{fig:gen_vad}
   \vspace{-0.4cm}
\end{figure}
Recent advancements in language models (LMs) have led to unprecedented common-sense ability and generalization performance across unseen data and tasks for the natural language processing domain~\cite{GPT3,chatgpt,Palm}.~The superior common-sense capability of LMs have led to the emergence of multi-modal LMs for diverse applications ranging from medical imaging to robotics~\cite{elixr,Palme,RT2}. However, incorporation of the reasoning ability of LMs into real-world autonomous driving tasks, to address generalization and long-tail scenarios, is yet to be fully-explored. 

To bridge this gap, we propose a \textbf{V}ision \textbf{L}anguage \textbf{P}lanning (VLP) framework, which integrates the common-sense capability of LMs into vision-based ADS for safe self-driving. Our VLP consists of two key components: Agent-centric Learning Paradigm (ALP) and Self-driving-car-centric Learning Paradigm (SLP), leveraging LMs to enhance the ADS from reasoning and decision-making aspects, respectively.

The BEV feature map serves as the source memory pool in ADS for downstream decoding tasks. It summarizes and encodes the driving environment surrounding the self-driving car, including vehicles, pedestrians, lanes, and more, into a unified feature map.~Hence, capturing comprehensive and necessary details in each local position of BEV is critical for safe and precise self-driving performance. To enhance the local semantic representation and reasoning capabilities of BEV, we introduce an innovative Agent-centric Learning Paradigm (ALP) module. ALP integrates the consistent feature space of a pretrained LM to revamp the agent features on the BEV, actively shaping semantics, and guiding the BEV reasoning process. Leveraging on the common sense and logic flow embedded in the LM, our ALP equips the ADS with robustness and consistent BEV feature space, enhancing its effectiveness in diverse driving scenarios.

In ADS, the planning module aggregates information from the preceding perception and prediction phases to make the final decisions for self-driving. This global perspective culminates in the formation of a planning query, directly influencing the safety and accuracy of the self-driving navigation.~Considering the critical role of the planning module within ADS, we also present a novel Self-driving-car-centric Learning Paradigm (SLP) to elevate the decoding and acquiring information ability of the planning query. In the SLP, we align the planning query with intended goals and the ego-vehicle driving status by leveraging the knowledge encoded in the pretrained LM. The language model's comprehension capabilities contribute to more informed decision-making during the planning phase as well as enabling a more robust planning query formation process.

Through VLP, we bridge the gap between human-like reasoning and autonomous driving, enhancing the model's contextual awareness and its ability to generalize effectively in complex, ever-changing real-world scenarios, as shown in Fig.\ref{fig:gen_vad}. The main contributions of this work are summarized as follows:
\begin{itemize}
    \item We propose VLP, a Vision Language Planning model, which incorporates reasoning capability of LMs into vision-based autonomous driving systems as an enhancement of motion planning and self-driving safety. 
    \item VLP is composed of novel components ALP and SLP, aiming to improve the ADS from self-driving BEV reasoning and self-driving decision-making aspects, respectively.
    \item Through extensive experiments in real-world driving scenarios, we show that VLP significantly and consistently outperforms the state-of-the-art vision-based approaches across a spectrum of driving tasks, including open-loop planning, multi-object tracking, motion forecasting, and more.
    \item We conduct the first new-city generalization study on the nuScenes dataset \cite{nuscenes} by training and testing on distinct cities, demonstrating the remarkable zero-shot generalization ability of our VLP approach over vision-only methods.
    \item To the best of our knowledge, this is the first work introducing LM into multiple stages of ADS to address the generalization ability in new cities and long-tail cases. 
\end{itemize}

%% file: sec/2_related.tex
\section{Related Work}
\label{sec:2_related}
\subsection{End-to-End Autonomous Driving}
\vspace{-0.1cm}

Early end-to-end approaches have adopted a single neural network to perform motion planning without explicitly designing intermediate tasks, such as prediction \cite{dave_nvidia}. Thus, such approaches not only suffer from the sub-optimal performance but also lack interpretability. Modular end-to-end frameworks introduce intermediate tasks for interpretability and provide improved performance \cite{PnPNet, P3,MP3, Vip3D,STP3,Uniad, VAD}.  
These interpretable end-to-end frameworks explicitly model perception, prediction and planning (P3) components and train them together with a joint optimization strategy to enhance planning. UniAD \cite{Uniad} and VAD \cite{VAD} are two prominent state-of-the-art end-to-end frameworks. UniAD leverages rasterized scene representations and explicitly identifies crucial components within the P3 framework. It coordinates all tasks through a unified query design for safe planning. On the other hand, VAD uses vectorized representations for efficient planning and safety improvement. While these approaches generally achieve good performance, their generalization performance have been limited due to lack of common sense and reasoning process.

\subsection{Vision-Language Models}
\vspace{-0.1cm}
Large language models (LLMs), which are trained on massive amounts of text data, have shown unprecedented success in language-related tasks \cite{GPT3,Palm,Llama}.~The capability of LLMs for common-sense understanding and versatility in tackling diverse language tasks has driven the exploration of multi-modal LLMs \cite{visualbert,clip,blip}.~Vision-language models (VLM) integrate visual and textual information to enable common sense understanding of the content by training on large scale open-world vision and text data~\cite{align,clip,blip,flamingo}. ~CLIP~\cite{clip} uses a contrastive learning objective to learn a joint embedding space for vision and text data, and has shown immense zero-shot generalization performance, showcasing the importance of VLMs compared to vision-only approaches. 

\subsection{Autonomous Driving with Language Models}
\vspace{-0.1cm}
In recent years, numerous works have been proposed for extending LMS to the embodied AI domain to improve zero-shot generalization performance \cite{yang2023foundation,RT2,Palme,voyager}. These works incorporate sensory inputs from embodied agents in the form of language with vision and language data for decision making process \cite{Palme, RT2}.~While these methods have mainly focused on the robotics domain, few works have been proposed for leveraging embodied LMs for autonomous driving tasks \cite{Dilu,GPTDriver,hasan2023vision,drivegpt4,lingo1,pan2024clip}. In particular, DiLu \cite{Dilu} and GPT-Driver \cite{GPTDriver} propose GPT-based driver agents for closed-loop simulation tasks.~In \cite{lingo1}, an open loop driving commentator, which combines vision and low-level driving actions with language, is proposed to interpret driving actions and for reasoning. Our work significantly differs from these approaches by leveraging LMs for end-to-end motion planning task for real-world driving applications. 

%% file: sec/3_model.tex
\section{Methodology}
\label{sec:3_model}
\vspace{-0.1cm}

\begin{figure*}[bt!]
  \centering\includegraphics[width=1.0\linewidth]{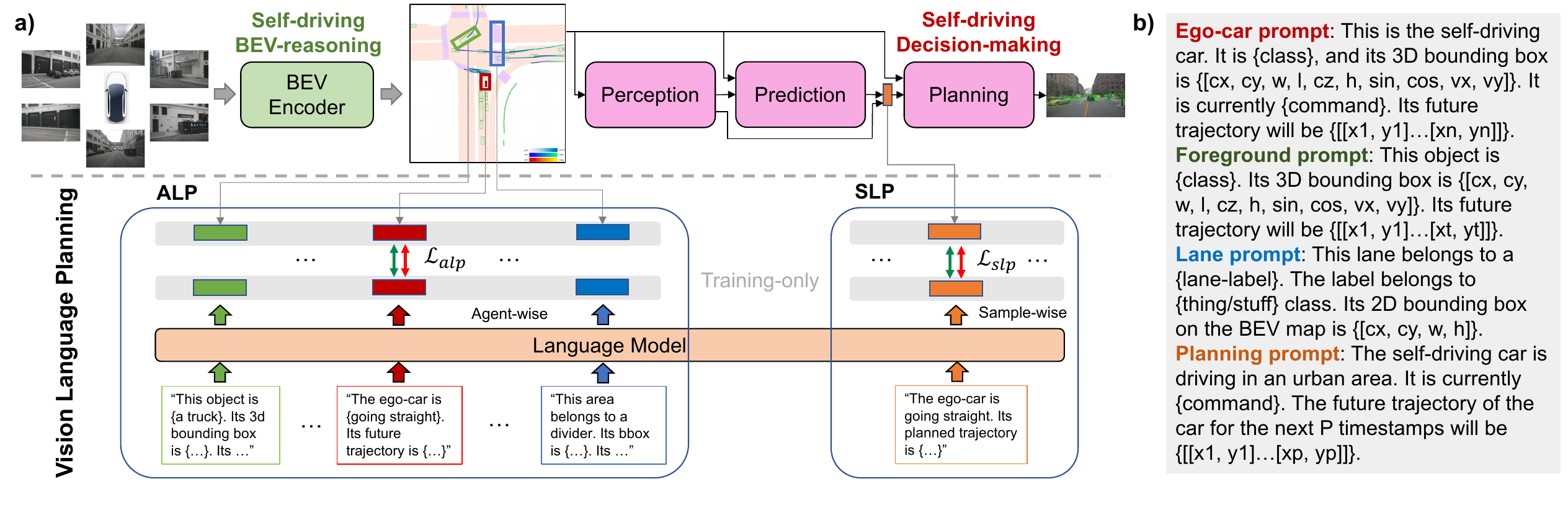}
  \vspace{-0.6cm}
   \caption{a) The overview of proposed vision language planning (VLP) framework. VLP enhances ADS from self-driving BEV-reasoning and self-driving decision-making aspects, through two innovative modules, ALP and SLP, respectively. Leveraging LM and contrastive learning, ALP conducts agent-wise learning for refining local details on BEV, while SLP engages sample-wise learning for advancing global context understanding ability of the ADS. VLP is only active during training, ensuring no additional parameters or computations are introduced during inference. b) Prompt formats used in VLP.}
   \label{fig:vlp}
   \vspace{-0.2cm}
\end{figure*}

We present a Vision Language Planning (VLP) model, which equips the ADS with the capacity to emulate human common sense and engage in contextual reasoning  for safe motion planning. Our proposed VLP model, illustrated in Fig. \ref{fig:vlp}a, comprises two innovative components that leverage LMs in both local and global contexts. The Agent-centric Learning Paradigm (ALP) concentrates on refining local details to enhance source memory reasoning, while the Self-driving-car-centric Learning Paradigm (SLP) focuses on guiding the planning process for the self-driving-car (SDC). Below, we first provide an overview of how general ADS work in Sec.\ref{ssec:Preliminary}. Subsequently, we describe our proposed ALP and SLP in detail in Sec.~\ref{ssec:bev-pro} and Sec.~\ref{ssec:ego-pro}, respectively.

\subsection{Preliminary}
\label{ssec:Preliminary}
\vspace{-0.1cm}

In vision-based ADS, a sequence of multi-view camera images serves as the source input, offering abundant visual data for downstream tasks. Initially, an image backbone extracts fundamental features for perception. These features are then processed by a BEV encoder to transform multi-view features into a unified 2D representation from a top-down view. This BEV feature map encapsulates crucial perception details like object positions, lane markings, and road boundaries~\cite{bevformer}. Acting as an information hub, this BEV feature map supports diverse downstream tasks such as 3D object tracking, mapping, motion prediction, occupancy prediction, and path planning~\cite{Uniad, VAD}. Specific task-related information is decoded using a query-based transformer decoder designed for each task to efficiently access the source memory \cite{Uniad}. In the final planning stage, an ego-query models the ego-vehicle's status and interacts with environmental features to determine the optimal path for autonomous driving.
\vspace{-0.1cm}

\subsection{Agent-centric Learning Paradigm: Enhancing BEV Source Memory} 
\vspace{-0.1cm}
\label{ssec:bev-pro}

The BEV feature map serves as the foundational source memory pool and holds a pivotal role in ADS. Ensuring that a BEV map provides comprehensive information with intricate details for driving is essential for enabling downstream decoders to make safe, precise, and human-like decisions.~However, the BEV in ADS is derived from multi-view camera images rather than a true bird's-eye-view image, which can introduce discrepancies between the produced BEV and the expected BEV representation. Therefore, to mitigate the discrepancies, we introduce the Agent-centric Learning Paradigm (ALP) to align the produced BEV with a true bird's-eye-view map. Through direct supervision on the BEV map with LM, our ALP enables the refinement of local details and alignment with the desired human perspective.

\noindent\textbf{Agent BEV features.} In our proposed ALP, three kinds of BEV agents are considered: ego-vehicle (self-driving-car), foreground (FG) objects, and lane elements. We first align the ground-truth area of each agent with the produced BEV map, and crop the regions of interest. We utilize the 3D bounding box to crop the ego-car and FG object area, and panoptic scene mask to segment the lane area. Subsequently, we perform a pooling operation on the obtained local BEV region, to generate a single feature representation for the corresponding agent.~After pooling, the local agent features in each sample along the batch are concatenated to formulate an Agent-BEV tensor denoted as $\mathcal{A}_{bev} \in \mathbb{R}^{N_{B} \times C}$, where $N_{B}$ and $C$ denote total number of agents in the batch and feature dimension, respectively.
The process can be formulated as:

\vspace{-0.4cm}
\begin{equation}
\small{
\begin{aligned}
\mathcal{A}_{bev}^{Ego} &= \text{Pool}(\text{Crop}(BEV, bbox^{Ego}_{3D})), \in \mathbb{R}^{C}, \\
\mathcal{A}_{bev}^{FG} &= \text{Pool}(\text{Crop}(BEV, bbox^{FG}_{3D})), \in \mathbb{R}^{C}, \\
\mathcal{A}_{bev}^{Lane} &= \text{Pool}(\text{Seg}(BEV, SegMask^{Lane})), \in \mathbb{R}^{C}, \\
\mathcal{A}_{bev} &= Batch([\mathcal{A}_{bev}^{Ego}; \mathcal{A}_{bev}^{FG}; \mathcal{A}_{bev}^{Lane}]), \in \mathbb{R}^{N_{batch} \times C}, \\
\label{eq:alp_bev}
\end{aligned}
}
\end{equation}

\vspace{-0.5cm}
where $Batch[;]$ denotes the concatenation operation, $\mathcal{A}_{bev}^{(.)}$ represents the single agent feature of the corresponding type, $bbox_{3D}^{(.)}$ and $SegMask^{Lane}$ indicate the ground-truth 3D bounding box and the lane segmentation mask, respectively. 

\noindent\textbf{Agent expectation features.} To ensure that local BEV features express the desired information, we conduct a BEV-expectation alignment process by leveraging LM and contrastive learning.~We precisely define the perceptual information expected from the corresponding agent, such as agent labels, bounding boxes, and future trajectories. These driving-related ground-truth information, which should also be embedded in the local BEV feature is formulated into a prompt as illustrated in Fig. \ref{fig:vlp}b. The description is then passed to the language encoder, $LM$, to generate the corresponding agent expectation feature. 
We apply an MLP layer $\mathcal{F}_{bev}$ to adapt the expectation feature to the BEV feature space. 
Then, the agent expectation features are concatenated along the batch to generate an Agent-Expectation tensor denoted as $\mathcal{A}_{exp} \in \mathbb{R}^{N_{B} \times C}$.
The procedure can be formulated as in Eq.~(\ref{eq:alp_prmpt}):

\vspace{-0.3cm}
\begin{equation}
\small{
\begin{aligned}
\mathcal{A}_{exp}^{Ego} &= \mathcal{F}_{bev}(LM(\boldsymbol{T}^{Ego}[y^{Ego}])), \in \mathbb{R}^{C}, \\
\mathcal{A}_{exp}^{FG} &= \mathcal{F}_{bev}(LM(\boldsymbol{T}^{FG}[y^{FG}])), \in \mathbb{R}^{C}, \\
\mathcal{A}_{exp}^{Lane} &= \mathcal{F}_{bev}(LM(\boldsymbol{T}^{Lane}[y^{Lane}])), \in \mathbb{R}^{C}, \\
\mathcal{A}_{exp} &= Batch([\mathcal{A}_{exp}^{ego}; \mathcal{A}_{exp}^{FG}; \mathcal{A}_{exp}^{lane}]),  
\label{eq:alp_prmpt}
\end{aligned}
}
\vspace{-0.2cm}
\end{equation}

where $\mathcal{A}_{exp}^{(.)}$, $\boldsymbol{T}^{(.)}$, and $y^{(.)}$ represent the single agent expectation feature, the description template, and the ground-truth in ADS for the respective agent. Note that during training, we freeze the $LM$ and only train the adaptation layer $\mathcal{F}_{bev}$ to save memory and retain the pre-trained knowledge in the $LM$.

\noindent\textbf{Contrastive Learning Formulation in ALP. } Given the agent BEV feature $\mathcal{A}_{bev}$ and the agent expectation feature $\mathcal{A}_{exp}$, the alignment between the produced BEV and the expected BEV is facilitated through a contrastive learning process~\cite{clip}.~In particular,
L2 normalization is applied to both $\mathcal{A}_{bev}$ and $\mathcal{A}_{exp}$ to standardize the feature vectors and ensure training stability. Then, matrix multiplication is performed between the normalized two-modality tensors with a learnable logit scale $\alpha_{alp}$ to produce a similarity matrix $\mathcal{S}_{pred} \in \mathbb{R}^{N_{B} \times N_{B}}$ in agent-wise. The ground truth for the similarity matrix, $\mathcal{S}_{gt}$, is a unit matrix, signifying that each agent's BEV feature should be closest to its corresponding expected feature in the shared space. To optimize this alignment, we apply a cross-entropy loss, $\mathcal{L}_{CE}$, along both the Agent-BEV mode axis and Agent-Expectation mode axis of the $\mathcal{S}_{pred}$, and compute an average value as the final ALP loss, $\mathcal{L}_{alp}$, for the BEV source reasoning part.
The entire process can be formally expressed as:
\begin{equation}
\vspace{-0.3cm}
\small{
\begin{aligned}
\mathcal{S}_{pred} &= \alpha_{alp} \times (\frac{\mathcal{A}_{bev}}{\|\mathcal{A}_{bev}\|_2} \otimes \frac{\mathcal{A}_{exp}}{\|\mathcal{A}_{exp}\|_2}), \\ %\in \mathbb{R}^{N_{B} \times N_{B}}, \\
\mathcal{L}_{alp} &= (\mathcal{L}_{CE}(\mathcal{S}_{pred}, \mathcal{S}_{gt}, dim=0) \\ &+ \mathcal{L}_{CE}(\mathcal{S}_{pred}, \mathcal{S}_{gt}, dim=1))/2. 
\label{eq:alp_loss}
\end{aligned}
}
\vspace{0.12cm}
\end{equation}

\subsection{SDC-centric Learning Paradigm: Enhancing Ego-vehicle Query}
\label{ssec:ego-pro}

\noindent\textbf{Ego-vehicle query feature.}~The ego-vehicle dynamic modeling is the core function of the ADS.
In previous ADS, a trainable ego-query $\mathcal{E}_{query}$ is applied to interact with other agents, $\mathcal{A}_{query}^{FG}$ and $\mathcal{A}_{query}^{Lane}$, on BEV map, to gather the self-driving perception/prediction information.~The produced ego-car query feature $\mathcal{E}_{qfeat}$ is further processed to predict the future waypoints of the self-driving car $y_{pred}^{plan}$. 
The process is formulated as in Eq.~(\ref{eq:slp_qfeat}):
\begin{equation}
\small{
\begin{aligned}
\mathcal{E}_{qfeat} &= \mathcal{M}^{inter}(\mathcal{E}_{query}, \mathcal{A}_{query}^{FG}, \mathcal{A}_{query}^{lane}) \in \mathbb{R}^{B \times C}, \\
y_{pred}^{plan} &= \mathcal{M}^{plan}(\mathcal{E}_{qfeat}) \in \mathbb{R}^{B \times P \times 2}, 
\label{eq:slp_qfeat}
\end{aligned}
}
\end{equation}
where $\mathcal{M}^{inter}$ and $\mathcal{M}^{plan}$ denote the intermediate ego information gathering module and the planning head in ADS, respectively. $P \times 2$ indicate the planned waypoints in the next P timesteps, and $B$ is the batch size during training. Although such mechanism can achieve good planning results, solely relying on numeric ground-truth can make it challenging to understand the rationale behind the ADS' decisions, which can result in inconsistent feature learning for the planning and lead to limited adaptability to new environments. To address these limitations, we propose a SDC-centric Learning Paradigm (SLP) to equip the ADS with the capability of making decisions from continuous and robust feature space. 

\noindent\textbf{Ego-vehicle planning feature.} We template a language description of the ego-vehicle status with the planning ground truth (GT) including high-level driving command and the future trajectory of the ego-vehicle, as illustrated in Fig. \ref{fig:vlp}b. The prompt is sent to the $LM$ to obtain the ground truth planning features of the ego-car embedding driving scenario information and human driving logic. 
An MLP layer $\mathcal{F}_{ego}$ is applied to adapt the textual planning features to the ego-query feature space. The process can be expressed as in Eq.~(\ref{eq:slp_prmpt}):

\vspace{-0.2cm}
\begin{equation}
\mathcal{E}_{prompt} = \mathcal{F}_{ego}(LM(\boldsymbol{T}_{slp}^{Ego}[y^{plan}])) \in \mathbb{R}^{B \times C},
\label{eq:slp_prmpt}
\end{equation}

\noindent where $\boldsymbol{T}_{slp}^{Ego}$ indicates the prompt for ego-car description used in SLP, and $y^{plan}$ denotes the planning ground truth. Note that the $LM$ is a shared off-the-shelf model for both ALP and SLP. As in ALP, only the adaptation layer $\mathcal{F}_{ego}$ is trainable during training to save memory and retain the pre-trained knowledge in $LM$.

\noindent\textbf{Contrastive Learning Formulation in SLP. } Similar to ALP, we employ a sample-wise contrastive learning approach for SLP to align the ego-vehicle query feature with the ego-vehicle textual planning feature as follows:

\vspace{-0.2cm}
\begin{equation}
\small{
\begin{aligned}
\mathcal{S}_{pred}^{slp} &= \alpha_{slp} \times (\frac{\mathcal{E}_{qfeat}}{\|\mathcal{E}_{qfeat}\|_2} \otimes \frac{\mathcal{E}_{prompt}}{\|\mathcal{E}_{prompt}\|_2}), \\
\mathcal{L}_{slp} &= (\mathcal{L}_{CE}(\mathcal{S}_{pred}^{slp}, \mathcal{S}_{gt}^{slp}, dim=0) \\ 
&+ \mathcal{L}_{CE}(\mathcal{S}_{pred}^{slp}, \mathcal{S}_{gt}^{slp}, dim=1))/2, 
\label{eq:slp_loss}
\end{aligned}
}
\end{equation}

where $\alpha_{slp}$ is a learnable logit scale, and $\mathcal{S}_{pred}^{slp}$ and $\mathcal{S}_{gt}^{slp}$ are the predicted and ground truth similarity matrices, respectively.
Aligning the two modes closely in the feature space, the contrastive process refines the produced ego-vehicle query using the ground truth-embedded linguistic feature. The sample-wise contrastive loss rectifies the relationships between individual samples from a human (linguistic) perspective, establishing a connection for the current data point with the world-wise common sense embedded in the pretrained language model.

\subsection{Training Loss}
The overall loss, $\mathcal{L}_{vlp}$, of our VLP training is composed of two parts: the BEV encoder reasoning loss $\mathcal{L}_{enc}$ and the decoder decision-making loss $\mathcal{L}_{dec}$ such that:

\vspace{-0.2cm}
\begin{equation}
\mathcal{L}_{vlp} = \omega_{enc} \mathcal{L}_{enc} + \omega_{dec}\mathcal{L}_{dec},
\label{eq:vllp_loss}
\end{equation}

where $\omega_{enc}$ and $\omega_{dec}$ represent the weights for $\mathcal{L}_{enc}$ and $\mathcal{L}_{dec}$, respectively.

The encoder reasoning loss, $\mathcal{L}_{enc}$, is equal to the loss produced in ALP module, i.e. 
$\mathcal{L}_{enc} = \mathcal{L}_{alp}$.
The decision-making loss includes the losses in all downstream tasks (perception/prediction/planning) in ADS and the SLP loss:

\vspace{-0.25cm}
\begin{equation}
\mathcal{L}_{dec} = \mathcal{L}_{perc} + \mathcal{L}_{pred}+ \mathcal{L}_{plan}+ \mathcal{L}_{slp}.
\label{eq:dec_loss}
\end{equation}

Note that the off-the-shelf LM head is discarded during inference, and thus, our method introduces no additional parameters and computations to the original ADS.

%% file: sec/4_exp.tex
\section{Experiments}
\subsection{Implementation Details}
\label{ssec:Implementation}
\textbf{Dataset.}~We conduct experiments on the challenging nuScenes dataset~\cite{nuscenes}, which is the first large-scale public dataset to provide data from the entire sensor suite of an autonomous vehicle (6 cameras, 1 LIDAR, 5 RADAR, GPS and IMU sensors). The nuScenes contains 1000 driving scenes from Boston and Singapore, two cities that are known for their dense traffic and highly challenging driving conditions. In our experiments, we utilize 6 camera images as our vision inputs.

\noindent\textbf{Baselines.}~We adopt two recent SOTA models in autonomous driving, namely UniAD~\cite{Uniad} and VAD~\cite{VAD}, as our baselines to evaluate the effectiveness of our approach.

\noindent\textbf{Training.}~We incorporate our proposed VLP training strategy into vision-only approaches of UniAD~\cite{Uniad} and VAD~\cite{VAD}, and refer to this enriched versions as VLP-UniAD and VLP-VAD, respectively, for clarity when reporting the results.~All methods are trained with the same hyper-parameters reported in the respective baselines~\cite{Uniad,VAD} for commensurate comparison. For language model, we use the pretrained LM in CLIP~\cite{clip}. In VLP study, we set T=6 (3 seconds) for planning, which is same as baselines \cite{Uniad} \cite{VAD}. Experiments are conducted with 8 NVIDIA Tesla A100 GPUs. More details regarding experimental setup can be found in supplementary materials.

\subsection{Open-loop Planning}\label{ssec:Planning}

Planning constitutes the cornerstone of any ADS, playing a pivotal role in ensuring safety and devising efficient routes for the ego-car. In Tab.~\ref{tab:planning}, we present a series of comparative experiments that showcase the performance of our open-loop planning in comparison to the baseline models. As can be seen in rows 4-6 of the table, the integration of just SLP leads to noticeable reductions in both the L2 error and collision rates for all the baseline models. Moving down the table, rows 7 to 9 demonstrate that the inclusion of both VLP components (SLP and ALP together) consistently yields further improvements in these planning metrics. In particular, VLP-UniAD shows a  28.1\% and 48.4\% reduction in terms of average L2 error and collision rate, respectively, compared to baseline UniAD. Similarly, in comparison with VAD, VLP-VAD achieves  35.9\% and 60.5\% reduction for average L2 error and collision rate, respectively. These significant results underscore the effectiveness of both SLP and ALP, as well as their adaptability across various ADS configurations. The reduced L2 error and collision rate achieved through VLP integration contribute to safer road planning in the realm of self-driving.

\begin{table*}[t]
\centering
\resizebox{0.7\linewidth}{!}{
    \begin{tabular}{l|l|cc|ccc|c|ccc|c}
        \toprule
        \multirow{2}{*}{ID} & \multirow{2}{*}{Model} & \multirow{2}{*}{SLP} & \multirow{2}{*}{ALP} & \multicolumn{4}{c|}{L2 (m) $\downarrow$} & \multicolumn{4}{c}{Col. Rate (\%) $\downarrow$} \\
         & & & & 1s & 2s & 3s & Avg. & 1s & 2s & 3s & Avg. \\
        \midrule
        0 & NMP~\cite{49_zeng2019end} & & & - & - & 2.31 & - & - & - & 1.92 \\
        1 & SA-NMP~\cite{49_zeng2019end} & & & - & - & 2.05 & - & - & - & 1.59 \\
        2 & FF~\cite{18_hu2021safe} & & & 0.55 & 1.20 & 2.54 & 1.43 & 0.06 & 0.17 & 1.07 & 0.43 \\
        3 & EO~\cite{24_khurana2022differentiable} & & & 0.67 & 1.36 & 2.78 & 1.60 & 0.04 & 0.09 & 0.88 & 0.33 \\
        4 & ST-P3~\cite{STP3} & & & 1.33 & 2.11 & 2.90 & 2.11 & 0.23 & 0.62 & 1.27 & 0.71 \\
        \midrule
        5 & UniAD~\cite{Uniad} & & & 0.48 & 0.96 & 1.65 & 1.03 & 0.05 & 0.17 & 0.71 & 0.31 \\
        \rowcolor{lightgray!80}
        6 & VLP-UniAD & $\checkmark$ & & \textbf{0.43} & \textbf{0.86} & \textbf{1.47} & \textbf{0.92} & \underline{\textbf{0.03}} & \textbf{0.15} & \textbf{0.48} & \textbf{0.22} \\
        \rowcolor{lightgray!80}
        7 & VLP-UniAD & $\checkmark$ & $\checkmark$ & \underline{\textbf{0.36}} & \underline{\textbf{0.68}} & \underline{\textbf{1.19}} & \underline{\textbf{0.74}} & \underline{\textbf{0.03}} & \underline{\textbf{0.12}} & \underline{\textbf{0.32}} & \underline{\textbf{0.16}} \\
        \midrule
        8 & VAD~\cite{VAD} & & & 0.46 & 0.76 & 1.12 & 0.78 & 0.21 & 0.35 & 0.58 & 0.38 \\
        \rowcolor{lightgray!80}
        9 & VLP-VAD & $\checkmark$ & & \underline{\textbf{0.26}} & \underline{\textbf{0.47}} & \underline{\textbf{0.78}} & \underline{\textbf{0.50}} & \textbf{0.12} & \textbf{0.17} & \textbf{0.42} & \textbf{0.23} \\
        \rowcolor{lightgray!80}
        10 & VLP-VAD & $\checkmark$  & $\checkmark$ & \textbf{0.30} & \textbf{0.53} & \textbf{0.84} & \textbf{0.55} & \underline{\textbf{0.01}} & \underline{\textbf{0.07}} & \underline{\textbf{0.38}} & \underline{\textbf{0.15}} \\
        \midrule
        \bottomrule
    \end{tabular}
}
\vspace{-0.2cm}
\caption{\textbf{Open-loop planning performance.} VLP achieves significant end-to-end planning performance improvement over counterpart vision only UniAD and VAD methods on the nuScenes validation dataset \cite{nuscenes}. Based on the planning results, we can conclude that both SLP and ALP components plays a vital role to ensure safe motion planning. }
\label{tab:planning}
\vspace{-0.2cm}
\end{table*}

\subsection{Perception and Prediction}
\label{ssec:perception}
\vspace{-0.1cm}
In this section, we showcase the consistent effectiveness of our proposed VLP across a spectrum of essential driving tasks. Our proposed approach excels in various perception and prediction tasks, including multi-object tracking, mapping, motion forecasting, occupancy prediction, 3D object detection, and vectorized scene segmentation. These results underscore the versatility and reliability of our VLP framework across a wide range of critical driving tasks.

\noindent\textbf{Multi-Object Tracking.} The results for multi-object tracking (MOT) are presented in Tab.~\ref{tab:mot}.~We apply the standard evaluation protocols of AMOTA (Average Multi-Object Tracking Accuracy), AMOTP (Average Multi-Object Tracking Precision), Recall, and IDS (Identity Switches) to evaluate the MOT performance.~As can be seen, with the incorporation of SLP, AMOTA, Recall, and IDS values all surpass those of UniAD, while AMOTP experiences a slight degradation. The combined integration of both SLP and ALP results in improvements across all metrics when compared to the UniAD baseline. 
The enhanced performance in MOT proves that our VLP can help the system better predict and respond to the movements of other objects on the road, reducing the risk of accidents.

\begin{table}[t]
\centering
\resizebox{1.0\linewidth}{!}{
    \begin{tabular}{l|cc|cccc}
        \toprule
        Model & SLP & ALP & AMOTA$\uparrow$ & AMOTP$\downarrow$ & Recall$\uparrow$ & IDS$\downarrow$ \\
        \midrule
        ViP3D~\cite{Vip3D} & & & 21.7 & 1.625 & 36.3 & - \\
        QD3DT~\cite{QD3DT} & & & 24.2 & 1.518 & 39.9 & - \\
        MUTR3D~\cite{MUTR3D} & & & 29.4 & 1.498 & 42.7 & 3822 \\
        \midrule
        UniAD~\cite{Uniad} & & & 35.9 & 1.320 & 46.7 & 906 \\
        \rowcolor{lightgray!80}
        VLP-UniAD & $\checkmark$ & & \textbf{36.6} & 1.332 & \textbf{46.8} & \textbf{820} \\
        \rowcolor{lightgray!80}
        VLP-UniAD & $\checkmark$ & $\checkmark$ & \underline{\textbf{36.8}} & \underline{\textbf{1.315}} & \underline{\textbf{47.3}} & \underline{\textbf{678}} \\
        \bottomrule
    \end{tabular}
}
\vspace{-0.2cm}
\caption{\textbf{Multi-object tracking.} VLP shows improved performance over vision-only MOT techniques.  }
\label{tab:mot}
\vspace{-0.2cm}
\end{table}

\noindent\textbf{Online Mapping.} In Tab.~\ref{tab:mapping}, we present the results of online mapping, encompassing four key mapping elements: lanes, drivable areas, dividers, and pedestrian crossings. The evaluation employs the Intersection-over-Union (IoU) metric to measure the overlap between the predicted and ground-truth maps.
The results reveal that the inclusion of the SLP leads to improvements in mapping accuracy for all four elements. Specifically, it enhances the IoU for drivable areas and crossings by 5.5\% and 6.5\%, respectively. Furthermore, the integration of both SLP and ALP yields even further improvement, with IoU achieving by 6.6\% and 10.2\% for drivable areas and crossings, respectively. These enhancements represent substantial gains compared to the SLP-only configuration.~These results underscore the valuable contributions of each paradigm in online mapping. They also demonstrate the effectiveness of VLP in bridging the gap between vision-based and language-based information, ultimately enhancing the system's comprehension of various road elements.

\begin{table}[t]
\centering
\resizebox{1.0\linewidth}{!}{
    \begin{tabular}{l|cc|cccc}
        \toprule
        Model & SLP & ALP & Lanes$\uparrow$ & Drivable$\uparrow$ & Divider$\uparrow$ & Crossing$\uparrow$ \\
        \midrule
        UniAD~\cite{Uniad} & & & 31.3 & 69.1 & 25.7 & 13.8 \\
        \midrule
        \rowcolor{lightgray!80}
        VLP-UniAD & $\checkmark$ & & \textbf{32.0} & \textbf{69.5} & \textbf{27.1} & \textbf{14.7} \\
        \rowcolor{lightgray!80}
        VLP-UniAD & $\checkmark$ & $\checkmark$ & \underline{\textbf{32.3}} & \underline{\textbf{70.2}} & \underline{\textbf{27.4}} & \underline{\textbf{15.2}} \\
        \bottomrule
    \end{tabular}
}
\vspace{-0.2cm}
\caption{\textbf{Online mapping.} VLP demonstrates improved segmentation IoU ($\%$) performance over the UniAD.   }
\label{tab:mapping}
\vspace{-0.2cm}
\end{table}

\noindent\textbf{Motion Forecasting.} We present the motion forecasting results in Tab.~\ref{tab:motion}, applying the same evaluation metrics as those in UniAD~\cite{Uniad}, namely minADE (minimum Average Displacement Error), minFDE (minimum Final Displacement Error), MR (Miss Rate), and EPA (End-to-end Prediction Accuracy).~The results demonstrate that the inclusion of the SLP contributes to a reduction in minADE, minFDE, and MR, while simultaneously increasing EPA for the ADS. This outcome signifies that the supervision and refinement applied to the final ego-car feature also has a positive influence on the motion prediction of other vehicles around the ego-car. The improvements are more obvious with the EPA metrics.~Equipped with full components of VLP, the performance of the ADS, in motion prediction, experiences further enhancements. This observation underscores the capability of ALP to empower the system in the identification and accurate prediction of the movements of various agents on the road.

\begin{table}[t]
\centering
\resizebox{1.0\linewidth}{!}{
    \begin{tabular}{l|cc|cccc}
        \toprule
        Model & SLP & ALP & minADE(m)$\downarrow$ & minFDE(m)$\downarrow$ & MR$\downarrow$ & EPA$\uparrow$ \\
        \midrule
        PnPNet~\cite{PnPNet} & & & 1.15 & 1.95 & 0.226 & 0.222 \\
        ViP3D~\cite{Vip3D} & & & 2.05 & 2.84 & 0.246 & 0.226 \\
        \midrule
        UniAD~\cite{Uniad} & & & 0.71 & 1.02 & 0.151 & 0.456 \\
        \rowcolor{lightgray!80}
        VLP-UniAD & $\checkmark$ & & 0.72 & 1.04 & 0.154 & 0.459 \\
        \rowcolor{lightgray!80}
        VLP-UniAD & $\checkmark$ & $\checkmark$ & \underline{\textbf{0.68}} & \underline{\textbf{0.98}} & \underline{\textbf{0.133}} & \underline{\textbf{0.460}} \\
        \midrule
        VAD~\cite{VAD} & & & 0.78 & 1.07 & 0.121 & 0.598 \\
        \rowcolor{lightgray!80}
        VLP-VAD & $\checkmark$ & & \underline{\textbf{0.77}} & \underline{\textbf{1.03}} & \underline{\textbf{0.110}} & \underline{\textbf{0.621}} \\
        \rowcolor{lightgray!80}
        VLP-VAD & $\checkmark$ & $\checkmark$ & \underline{\textbf{0.77}} & \textbf{1.05} & \textbf{0.115} & \underline{\textbf{0.621}} \\
        \bottomrule
    \end{tabular}
}
\vspace{-0.2cm}
\caption{\textbf{Motion forecasting.} VLP achieves better motion forecasting over counterpart vision based methods.}
\label{tab:motion}
\vspace{-0.2cm}
\end{table}

\noindent\textbf{Occupancy Prediction.} We present the occupancy prediction results in Tab.~\ref{tab:occ}, which have been obtained using the  IoU and Video Panoptic Quality (VPQ) within two distance ranges around the ego-car, namely near ("-n."), covering a $30 \times 30m$ area, and far ("-f."), spanning a $50 \times 50m$ area. The results highlight the effectiveness of the SLP in consistently improving all four metrics. SLP, primarily focusing on enhancing the ego-car feature, demonstrates consistent performance gains in occupancy prediction. The inclusion of ALP with SLP shows similar improvement over the baseline as well. Since both configurations of the VLP prioritize attention to the ego-car during training, more improvements are observed within the nearby areas.

\begin{table}[t]
\centering
\resizebox{1.0\linewidth}{!}{
    \begin{tabular}{l|cc|cccc}
        \toprule
        Model & SLP & ALP & IoU-n.$\uparrow$ & IoU-f.$\uparrow$ & VPQ-n.$\uparrow$ & VPQ-f.$\uparrow$ \\
        \midrule
        FIERY~\cite{fiery} & & & 59.4 & 36.7 & 50.2 & 29.9 \\
        StretchBEV~\cite{stretchbev} & & & 55.5 & 37.1 & 46.0 & 29.0 \\
        ST-P3~\cite{STP3} & & & - & 38.9 & - & 32.1 \\
        \midrule
        UniAD~\cite{Uniad} & & & 63.4 & 40.2 & 54.7 & 33.5 \\
        \rowcolor{lightgray!80}
        VLP-UniAD & $\checkmark$ & & \underline{\textbf{64.2}} & \underline{\textbf{40.7}} & \textbf{55.8} & \underline{\textbf{34.5}} \\
        \rowcolor{lightgray!80}
        VLP-UniAD & $\checkmark$ & $\checkmark$ & \textbf{64.1} & 40.2 & \underline{\textbf{55.9}} & \textbf{34.1} \\
        \bottomrule
    \end{tabular}
}
\vspace{-0.2cm}
\caption{\textbf{Occupancy prediction.} VLP shows improvement in both near ($30\times30 m$) and far ($50\times50 m$) ranges, denoted as ``n." and ``f.", respectively.}
\label{tab:occ}
\vspace{-0.2cm}
\end{table}

\noindent\textbf{3D Object Detection.} 
We show the 3D object detection results in Tab.~\ref{tab:det} using nuScenes detection metrics of mean Average Precision (mAP), Average Translation Error (ATE), Average Scale Error (ASE), and nuScenes detection score (NDS) as indicated in \cite{nuscenes}.~As observed in the table, the incorporation of the VLP consistently leads to improvements in NDS. Notably, models incorporating both the SLP and ALP generally perform better than those with SLP alone, providing empirical evidence of the ALP's effectiveness in enhancing the local BEV representation ability.

\begin{table}[t]
\centering
\resizebox{1.0\linewidth}{!}{
    \begin{tabular}{l|cc|cccc}
        \toprule
        Model & SLP & ALP & mAP$\uparrow$ & mATE$\downarrow$ & mASE$\downarrow$ & NDS$\uparrow$ \\
        \midrule
        VAD~\cite{VAD} & & & 0.27 & 0.70 & 0.30 & 0.389 \\
        \midrule
        \rowcolor{lightgray!80}
        VLP-VAD & $\checkmark$ & & 0.27 & \underline{\textbf{0.67}} & 0.30 & \textbf{0.394} \\
        \rowcolor{lightgray!80}
        VLP-VAD & $\checkmark$ & $\checkmark$ & \underline{\textbf{0.28}} & \underline{\textbf{0.67}} & 0.30 & \underline{\textbf{0.406}} \\
        \bottomrule
    \end{tabular}
}
\vspace{-0.2cm}
\caption{\textbf{3D object detection.} VLP achieves improved object detection performance over the baseline.  }
\label{tab:det}
\vspace{-0.2cm}
\end{table}

\noindent\textbf{Vectorized Scene Segmentation in VAD.} The VAD~\cite{VAD} framework represents the driving scene as a fully vectorized structure, categorizing map elements into road boundaries, dividers, and pedestrian crossings. Thus, vectorized scene segmentation experiments are conducted within the VAD framework, with IoU serving as the evaluation metric.~As depicted in Tab.~\ref{tab:vec-seg}, the introduction of SLP yields substantial improvements. More specifically, the segmentation for boundaries, dividers, and crossings sees improvements of 14.9\%, 19.2\%, and 23.1\%, respectively, contributing to an overall increase of 18.6\% in the mean IoU (mIoU). VLP with SLP and ALP also achieves similar improvement over the VAD baseline. This increased accuracy is crucial for developing a safe and efficient self-driving system.

\begin{table}[t]
\centering
\resizebox{1.0\linewidth}{!}{
    \begin{tabular}{l|cc|cccc}
        \toprule
        Model & SLP & ALP & Boundary$\uparrow$ & Divider$\uparrow$ & Crossing$\uparrow$ & mIoU$\uparrow$ \\
        \midrule
        VAD~\cite{VAD} & & & 45.6 & 42.2 & 31.6 & 39.8 \\
        \midrule
        \rowcolor{lightgray!80}
        VLP-VAD & $\checkmark$ & & \underline{\textbf{52.4}} & \underline{\textbf{50.3}} & \textbf{38.9} & \underline{\textbf{47.2}} \\
        \rowcolor{lightgray!80}
        VLP-VAD & $\checkmark$ & $\checkmark$ & \textbf{49.4}  & \textbf{48.4} & \underline{\textbf{39.2}} & \textbf{45.7} \\
        \bottomrule
    \end{tabular}
}
\vspace{-0.2cm}
\caption{\textbf{Vectorized scene segmentation.} VLP shows improved segmentation IoU ($\%$) performance over the VAD.}
\label{tab:vec-seg}
\vspace{-0.2cm}
\end{table}

\subsection{Generalization}
\label{ssec:generalization}
\vspace{-0.1cm}

Autonomous vehicles are intended to operate in diverse environments.~Generalization allows a system to be
deployed in various urban landscapes, suburban areas, or even rural settings, making the technology applicable and accessible on a broader scale. Long-tail cases, which are scenarios that occur infrequently, are often underrepresented in training data.~Generalization helps the system cope with these rare but critical situations, reducing the risk of biased decision-making based on inadequate exposure during training.~In this section, we evaluate the VLP in terms of generalization ability to new cities and long-tail cases.

\noindent\textbf{New-city Generalization.} 
To keep the same number of multi-view camera inputs, we construct a multi-city dataset exclusively from nuScenes, by encompassing data from two distinct urban environments, Boston and Singapore.
To comprehensively assess generalizability, we conduct two sets of experiments: (i) training on Boston and testing on Singapore; and (ii) conversely, training on Singapore and testing on Boston.~The results, presented in Tab.~\ref{tab:gen-new-city}, reveal a noteworthy reduction in planning L2 error and collision rates for the baselines in both scenarios with the integration of VLP. In particular, VLP-VAD achieves strong generalization over counterpart vision-only VAD approach (in Boston: 15.1\% and 18.5\%, in Singapore: 19.2\% and 48.7\% reduction in terms of average L2 error and collision rates, respectively).
This signifies VLP's prowess in enhancing the safety and reliability of the ADS as well as its capacity to transcend training confines and excel in diverse real-world conditions.

\begin{table}[t]
\centering
\resizebox{1.0\linewidth}{!}{
    \begin{tabular}{l|c|cc|cc}
        \toprule
        \multirow{2}{*}{Model} & \multirow{2}{*}{VLP} & \multicolumn{2}{c|}{Boston} & \multicolumn{2}{c}{Singapore} \\
         & & Avg.L2$\downarrow$ & Avg.Col$\downarrow$ & Avg.L2$\downarrow$ & Avg.Col$\downarrow$ \\
        \midrule
        UniAD~\cite{Uniad} & & 1.24 & 0.32 & 1.05 & 0.37 \\
        \rowcolor{lightgray!80}
        VLP-UniAD & $\checkmark$ & \textbf{1.14} & \textbf{0.26} & \textbf{0.87} & \textbf{0.34} \\        
        % VAD-base & & 0.91 & 0.71 & 0.74 & 0.44 \\
        \midrule
        VAD~\cite{VAD} & & 0.86 & 0.27 & 0.78 & 0.39 \\
        \rowcolor{lightgray!80}
        VLP-VAD & $\checkmark$ & \textbf{0.73} & \textbf{0.22} & \textbf{0.63} & \textbf{0.20} \\
        % \rowcolor{lightgray!80}
        % VLP-VAD-base & $\checkmark$ & 0.81 & 0.45 & 0.67 & 0.38 \\
        \bottomrule
    \end{tabular}
}
\vspace{-0.2cm}
\caption{\textbf{New-city generalization.} To evaluate the generalization performance, we train the model on Boston and test it on Singapore, and vice versa. VLP shows remarkable zero-shot generalization performance improvement over the vision-only methods.}
\label{tab:gen-new-city}
\vspace{-0.2cm}
\end{table}

\noindent\textbf{Long-tail Generalization.} 
We assess the performance of UniAD in Multi-Object Tracking (MOT) and VAD in 3D detection concerning long-tail scenarios, as detailed in Tab.~\ref{tab:mot-long-tail} and Tab.~\ref{tab:det-long-tail-od}, respectively. Utilizing the long-tail split methodology from \cite{longTail}, derived from per-class object counts, we observe that VLP consistently enhances the generalization capabilities of both frameworks.~Specifically, VLP provides substantial improvements, increasing AMOTA and Recall by 3.7\% and 7.4\%, respectively, for UniAD. In the case of VAD, VLP improves the mean Average Precision (mAP)  by 15.9\%. These results highlights the robust feature space cultivated by our proposed VLP, emphasizing its efficacy in handling long-tail scenarios.

\begin{table}[t]
\begin{subtable}{0.5\textwidth}
\centering
\resizebox{1.0\linewidth}{!}{
    \begin{tabular}{l|c|ccccc|}
        \toprule
        Model & VLP & AMOTA$\uparrow$ & AMOTP$\downarrow$ & Recall$\uparrow$ & IDS$\downarrow$ \\
        \midrule
        UniAD\cite{Uniad} & & 29.6 & 1.446 & 39.2 & 68 \\
        \midrule
        \rowcolor{lightgray!80}
        VLP-UniAD & $\checkmark$ & \textbf{30.7} & \textbf{1.435} & \textbf{42.1} & \textbf{66} \\
        \bottomrule
    \end{tabular}
}
\caption{Multi-object tracking on long-tail cases.}
\label{tab:mot-long-tail}
\end{subtable}

\medskip
\begin{subtable}{0.5\textwidth}

\centering
\resizebox{1.0\linewidth}{!}{
    \begin{tabular}{l|c|cccc}
        \toprule
        Model & VLP & mAP$\uparrow$ & mATE$\downarrow$ & mASE$\downarrow$ & mAOE$\downarrow$ \\
        \midrule
        VAD~\cite{VAD} & & 17.6 & 0.79 & 0.33 & 0.95 \\
        \midrule
        \rowcolor{lightgray!80}
        VLP-VAD & $\checkmark$ & \textbf{20.4} & \textbf{0.75} & 0.33 & \textbf{0.83} \\
        \bottomrule
    \end{tabular}
}
\caption{Object detection on long-tail cases.}
\label{tab:det-long-tail-od}
\end{subtable}
\vspace{-0.5cm}
\caption{\textbf{Long-tail generalization.} VLP shows consistent improvement for long-tail cases accross both a) multi-object tracking and b) object detection tasks. }
\label{tab:det-long-tail}
\vspace{-0.1cm}
\end{table}

\subsection{Ablation Studies}
\label{ssec:ablation}
\textbf{Effectiveness of SLP and ALP.} VLP comprises ALP and SLP components. Extensive experiments presented in prior tables show the importance and necessity of both SLP and ALP for enhanced planning and safety. 

\noindent\textbf{Prompt Format.}
The design of the prompt format is pivotal for the success of our VLP, serving as the structured input that guides the learning process. We have performed ablation studies to systematically investigate the impact of different components of the ground truth information included in the prompt. The driving ground truth encompasses labels, bounding boxes, trajectories, and high-level commands. To demonstrate the indispensability of including all aspects of the ground truth in the prompt, we conducted experiments where each element was selectively removed in individual settings. The results, presented in Tab.~\ref{tab:abl-prmpt}, clearly indicate that the exclusion of any ground truth component leads to a degradation in planning ability. This underscores the necessity of incorporating the entire spectrum of ground truth information in the prompt for optimal performance.

\begin{table}[t]
\vspace{-0.2cm}
\centering
\resizebox{0.9\linewidth}{!}{
    \begin{tabular}{cccc|cc}
        \toprule
        label-gt & bbox-gt & traj-gt & commd-gt & Avg.L2$\downarrow$ & Avg.Col$\downarrow$ \\
        \midrule
          \xmark &&&  & 0.64 & 0.23 \\
          & \xmark && & 0.56 & 0.30 \\
          && \xmark & & 0.59 & 0.36 \\
          &&& \xmark  & 0.59 & 0.26 \\
        \midrule
        $\checkmark$ & $\checkmark$ & $\checkmark$ & $\checkmark$ & \textbf{0.52} & \textbf{0.17} \\
        \bottomrule
    \end{tabular}
}
\vspace{-0.2cm}
\caption{\textbf{Ablation for prompt information.} All ground-truth information components contributes to improved planning performance. We provide full task ablation results in the supplementary.}
\label{tab:abl-prmpt}
\vspace{-0.2cm}
\end{table}

%% file: sec/5_conclusion.tex
\section{Conclusion}
\label{sec:5_con}

We have introduced a novel Vision-Language-Planning (VLP) approach to enhance the capabilities of Autonomous Driving Systems (ADS). Our approach leverages both self-driving-car-centric learning paradigm (SLP) and agent-wise learning paradigm (ALP) guided by language prompts to create a comprehensive understanding of the environment. Through a series of experiments on various driving tasks, we have demonstrated the effectiveness of our VLP approach in improving perception, prediction, and planning aspects of ADS.
The generalization experiments showcased the robustness of our VLP approach, proving its adaptability to new cities and long-tail cases. By extending the capabilities of ADS beyond the training environment, our VLP approach paves the way for safer and more reliable autonomous driving in real-world conditions.

\noindent\textbf{Limitations.} Our experiments are currently confined to the nuScenes dataset and camera modality as baseline vision-based approaches. We will assess VLP on a broader range of datasets and sensor modalities in our future work. 

%% file: sec/X_suppl.tex
\clearpage
\setcounter{page}{1}
\maketitlesupplementary

\section{Experiment Setup}
\label{sec:sup-setting}
\textbf{Model Components.}
UniAD is composed of a BEV extractor followed by five transformer decoder-based P3 modules for 3D object tracking, mapping, motion forecasting, occupancy prediction, and planning, respectively.
VAD is composed of a BEV extractor followed by four transformer decoder-based modules for 3D object detection, scene segmentation, motion forecasting, and planning, respectively.

\noindent\textbf{Model Hyperparameters.}
We provide the detailed training configurations for UniAD-based and VAD-based models in the Tab.~\ref{tab:train-recipe}.

\begin{table*}[t]
\centering
\resizebox{0.8\linewidth}{!}{
    \begin{tabular}{l|cc}
        \toprule
        Configs & UniAD-based & VAD-based \\
        \midrule
        point cloud range & $[-51.2, -51.2, -5.0, 51.2, 51.2, 3.0]$ & $[-15.0, -30.0, -2.0, 15.0, 30.0, 2.0]$ \\
        transformer decoder dimension & 256 & 256 \\
        BEV size & $200\times200$ & $100\times100$ \\
        queue length & 3 & 3 \\
        motion predict steps & 12 & 6 \\
        motion predict modes & 6 & 6 \\
        planning steps & 6 & 6 \\
        backbone & RN101 & RN50 \\
        optimizer & AdamW & AdamW \\
        learning rate & 2e-4 & 2e-4 \\
        weight decay & 0.01 & 0.01 \\
        epoch & 20 & 60 \\
        batch size & 8 & 8 \\
        \bottomrule
    \end{tabular}
}
\caption{\textbf{Training configurations for UniAD-based and VAD-based models.}}
\label{tab:train-recipe}
\end{table*}

\section{Ablation Study}
\label{sec:sup-abl}

\textbf{Prompt Information.} We present ablation studies on prompt information for UniAD and VAD, accompanied by downstream task results in Tab.~\ref{tab:sup-abl-prmpt-uniad} and \ref{tab:sup-abl-prmpt-vad}. Notably, training with completed ground truth information embedded in the prompts consistently yields superior performance across all downstream tasks for both UniAD and VAD. In comparison to the baseline (1st line in each table), models equipped with VLP, regardless of the ground truth information included, consistently outperform the model without VLP. These observations underscore the efficacy of the proposed VLP in enhancing model performance.

\noindent \textbf{Different LMs.} We explored several pretrained LMs for integration with our VLP. As shown in Tab.~\ref{tab:various-lms}, each integration led to consistent improvements in decision-making over two baselines, proving the effectiveness of VLP design (leverages LM locally and globally). Also, it underscores our contribution of integrating language understanding into visual models for autonomous driving, irrespective of the specific choice of the LM. Our choice was driven by a balance between computational efficiency and effectiveness. Integration with LLAMA will be our future work.

\noindent \textbf{ALP and SLP.} As shown in Tab.~\ref{tab:ALP-only}, both SLP and ALP consistently improve all tasks.~SLP focuses more on the final planning as it works for the ego-car query while ALP works for the shared BEV map.~When ALP and SLP operate concurrently, the model puts more efforts on optimizing the final planning.~While there is a relatively smaller margin of improvement in scene segmentation and occupancy prediction, the VLP still significantly surpasses the baseline (+VLP: +5.9\%). Trajectory planning is the most important step which guarantees the safety and efficiency of ADS. Improvement on it is the main target of VLP.

\begin{table}[b!]
\centering
\vspace{-0.6cm}
\resizebox{1.0\linewidth}{!}{
    \begin{tabular}{c|cc|c|cc}
        \toprule
        \multirow{2}{*}{Pretrained LM} & \multicolumn{2}{c|}{Planning} & \multirow{2}{*}{Pretrained LM} & \multicolumn{2}{c}{Planning} \\
         & avg.L2 $\downarrow$ & avg.Col $\downarrow$ & & avg.L2 $\downarrow$ & avg.Col $\downarrow$ \\
        \midrule
        \rowcolor{lightgray!45} \multicolumn{3}{c|}{UniAD} & \multicolumn{3}{c}{VAD} \\
        \midrule
         - & 1.03 & 0.31 & - & 0.82 & 0.93 \\
        \midrule
        \rowcolor{green!25} \multicolumn{3}{c|}{+VLP} & \multicolumn{3}{c}{+VLP} \\
        \midrule
        GPT2 & 0.75 (+27.2\%) & 0.16 (+48.4\%) & GPT2 & 0.61 (+25.6\%) & 0.29 (+68.8\%) \\
        CLIP-RN50x64-LM & 0.74 (+28.2\%) & 0.16 (+48.4\%) & CLIP-RN50x64-LM & 0.62 (+24.4\%) & 0.30 (+67.7\%) \\
        CLIP-RN101-LM & 0.76 (+26.2\%) & 0.21 (+32.3\%) & CLIP-RN101-LM & 0.57 (+30.5\%) & 0.35 (+62.4\%) \\
        CLIP-ViT-L/14-336px-LM & 0.73 (+29.1\%) & 0.24 (+22.6\%) & CLIP-ViT-L/14-336px-LM & 0.64 (+22.0\%) & 0.45 (+51.6\%) \\
        \bottomrule
    \end{tabular}
}
\vspace{-0.3cm}
\caption{Open-loop planning results with various LMs.}
\label{tab:various-lms}
 \vspace{-0.4cm}
\end{table}
\begin{table}[b]
\centering
\resizebox{1.0\linewidth}{!}{
    \begin{tabular}{l|c|c|cc|cc}
        \toprule
        \multirow{2}{*}{VLP} & Vectorized Scene Seg. & Occupancy Pred. & \multicolumn{2}{c|}{Plan Val.} & \multicolumn{2}{c}{Plan Singapore} \\
         & mIoU$\uparrow$ & IoU-n.$\uparrow$ IoU-f.$\uparrow$ VPQ-n.$\uparrow$ VPQ-f.$\uparrow$ & avg.L2 $\downarrow$ & avg.Col $\downarrow$ & avg.L2 $\downarrow$ & avg.Col $\downarrow$ \\
        \midrule
        - & 39.8 & 63.4 \quad 40.2 \quad 54.7 \quad 33.5 & 0.78 & 0.38 & 0.78 & 0.39 \\
    +SLP & 47.2 & 64.2 \quad 40.7 \quad 55.8 \quad 34.5 & 0.50 & 0.23 & 0.66 & 0.25 \\
    +ALP & 47.6 & 64.5 \quad 41.0 \quad 56.1 \quad 34.7 & 0.52 & 0.26 & 0.68 & 0.28 \\
    +VLP & 45.7 & 64.1 \quad 40.2 \quad 55.9 \quad 34.1 & 0.55 & 0.15 & 0.63 & 0.20 \\
        \bottomrule
    \end{tabular}
}
\vspace{-0.3cm}
\caption{Ablation study on each component of VLP.}
\label{tab:ALP-only}
\end{table}

\begin{table}[t]
\centering
\resizebox{1.0\linewidth}{!}{
    \begin{tabular}{l|cc|cc|ccccc}
        \toprule
        Model & SLP & ALP & mAP$\uparrow$ & NDS$\uparrow$ & mATE$\downarrow$ & mASE$\downarrow$ & mAOE$\downarrow$ & mAVE$\downarrow$ & mAAE$\downarrow$ \\
        \midrule
        VAD~\cite{VAD} & & & 17.6 & 29.9 & 0.79 & 0.33 & 0.95 & 0.65 & \underline{\textbf{0.17}} \\
        \midrule
        \rowcolor{lightgray!80}
        VLP-VAD & $\checkmark$ & & \textbf{19.1} & \textbf{31.2} & \textbf{0.76} & 0.33 & 1.00 & \underline{\textbf{0.50}} & 0.25 \\
        \rowcolor{lightgray!80}
        VLP-VAD & $\checkmark$ & $\checkmark$ & \underline{\textbf{20.4}} & \underline{\textbf{33.2}} & \underline{\textbf{0.75}} & 0.33 & \underline{\textbf{0.83}} & \textbf{0.60} & 0.19 \\
        \bottomrule
    \end{tabular}
}
\caption{\textbf{3D Object detection in challenging long-tail scenarios.}}
\label{tab:sup-long-tail-3d}
\end{table}

\begin{table*}[t]
\centering
\resizebox{1.0\linewidth}{!}{
    \begin{tabular}{cccc|ccc|cc|ccc|cccc|cc}
        \toprule
        \multicolumn{4}{c|}{Included Ground Truth} & \multicolumn{3}{c|}{Tracking} & \multicolumn{2}{c|}{Mapping} & \multicolumn{3}{c|}{Motion Forecasting} & \multicolumn{4}{c|}{Occupancy Prediction} & \multicolumn{2}{c}{Planning} \\
        Label & Bbox & Trajectory & Command & AMOTA↑ & AMOTP↓ & IDS↓ & IoU-lane↑ & IoU-road↑ & minADE↓ & minFDE↓ & MR↓ & IoU-n.↑ & IoU-f.↑ & VPQ-n.↑ & VPQ-f.↑ & avg.L2↓ & avg.Col.↓ \\
        \midrule
        \xmark & \xmark & \xmark & \xmark & 35.2 & 1.353 & 720 & 29.5 & 66.7 & 0.83 & 1.24 & 0.187 & 59.5 & 38.5 & 49.9 & 29.4 & 0.93 & 0.38 \\
        
        \midrule
        \xmark & $\checkmark$ & $\checkmark$ & $\checkmark$ & 35.5 & 1.331 & 670 & 30.8 & 66.8 & 0.78 & 1.23 & 0.180 & 59.8 & 39.1 & 51.2 & 30.7 & 0.87 & 0.31 \\
        $\checkmark$ & \xmark & $\checkmark$ & $\checkmark$ & 35.9 & 1.331 & 650 & 31.2 & 67.5 & 0.75 & 1.19 & 0.162 & 61.9 & 38.5 & 53.1 & 31.4 & 0.82 & 0.32 \\
        $\checkmark$ & $\checkmark$ & \xmark & $\checkmark$ & 35.4 & 1.340 & 710 & 31.1 & 67.2 & 0.81 & 1.20 & 0.179 & 62.1 & 38.7 & 53.8 & 31.5 & 0.89 & 0.36 \\
        $\checkmark$ & $\checkmark$ & $\checkmark$ & \xmark & 36.1 & 1.329 & 610 & 31.6 & 68.9 & 0.76 & 1.12 & 0.168 & 61.7 & 38.9 & 54.1 & 31.9 & 0.80 & 0.29 \\
        
        \midrule
        $\checkmark$ & $\checkmark$ & $\checkmark$ & $\checkmark$ & 36.2 & 1.320 & 620 & 31.7 & 69.1 & 0.72 & 1.11 & 0.156 & 62.4 & 39.3 & 54.2 & 32.9 & 0.78 & 0.24 \\
        \bottomrule
    \end{tabular}
}
\caption{\textbf{Detailed ablations on the effectiveness of prompt information with UniAD-based models.} Our ablation studies on prompt information for UniAD reveal a significant performance boost when training with completed ground truth information embedded in the prompts.}
\label{tab:sup-abl-prmpt-uniad}
\end{table*}

\begin{table*}[t]
\centering
\resizebox{1.0\linewidth}{!}{
    \begin{tabular}{cccc|cc|cccc|cccc|cccc|cc}
        \toprule
        \multicolumn{4}{c|}{Included Ground Truth} & \multicolumn{2}{c|}{3D Object Detection} & \multicolumn{4}{c|}{Vectorized Scene Segmentation} & \multicolumn{4}{c|}{Motion Forecasting} & \multicolumn{2}{c}{Planning} \\
        Label & Bbox & Trajectory & Command & mAP↑ & NDS↑ & Boundary↑ & Divider↑ & Crossing↑ & mIoU↑ & minADE↓ & minFDE↓ & MR↓ & EPA↑ & avg.L2↓ & avg.Col.↓ \\
        \midrule
        \xmark & \xmark & \xmark & \xmark & 22.6 & 33.3 & 43.3 & 43.6 & 34.0 & 40.3 & 0.86 & 1.20 & 0.143 & 0.526 & 0.98 & 0.80 \\
        \midrule
        \xmark & $\checkmark$ & $\checkmark$ & $\checkmark$ & 24.9 & 35.7 & 45.2 & 42.5 & 32.9 & 40.9 & 0.76 & 1.06 & 0.127 & 0.522 & 0.64 & 0.23 \\
        $\checkmark$ & \xmark & $\checkmark$ & $\checkmark$ & 25.7 & 37.6 & 46.5 & 44.4 & 30.4 & 40.4 & 0.78 & 1.05 & 0.119 & 0.549 & 0.56 & 0.30 \\
        $\checkmark$ & $\checkmark$ & \xmark & $\checkmark$ & 27.0 & 38.7 & 44.9 & 40.0 & 34.8 & 40.0 & 0.86 & 1.19 & 0.141 & 0.532 & 0.59 & 0.36 \\
        $\checkmark$ & $\checkmark$ & $\checkmark$ & \xmark & 26.6 & 38.0 & 44.0 & 40.3 & 31.4 & 38.9 & 0.79 & 1.06 & 0.121 & 0.551 & 0.59 & 0.26 \\
        \midrule
        $\checkmark$ & $\checkmark$ & $\checkmark$ & $\checkmark$ & 27.3 & 39.0 & 47.1 & 45.3 & 36.0 & 42.8 & 0.77 & 1.05 & 0.117 & 0.551 & 0.52 & 0.17 \\
        \bottomrule
    \end{tabular}
}
\caption{\textbf{Detailed ablations on the effectiveness of prompt information with VAD-based models.} Our ablation studies on prompt information for VAD shows a considerable performance improvement when training with completed ground truth information embedded in the prompts.}
\label{tab:sup-abl-prmpt-vad}
\end{table*}

\begin{table*}[t]
\centering
\resizebox{1.0\linewidth}{!}{
    \begin{tabular}{c|ccc|cc|ccc|cccc|cc}
        \toprule
        \multirow{2}{*}{Pretrained LM} & \multicolumn{3}{c|}{Tracking} & \multicolumn{2}{c|}{Mapping} & \multicolumn{3}{c|}{Motion Forecasting} & \multicolumn{4}{c|}{Occupancy Prediction} & \multicolumn{2}{c}{Planning} \\
          & AMOTA↑ & AMOTP↓ & IDS↓ & IoU-lane↑ & IoU-road↑ & minADE↓ & minFDE↓ & MR↓ & IoU-n.↑ & IoU-f.↑ & VPQ-n.↑ & VPQ-f.↑ & avg.L2↓ & avg.Col.↓ \\
        \midrule
        - & 35.2 & 1.353 & 720 & 29.5 & 66.7 & 0.83 & 1.24 & 0.187 & 59.5 & 38.5 & 49.9 & 29.4 & 0.93 & 0.38 \\
        \midrule
        CLIP-RN50x64-LM & 32.9 & 1.384 & 780 & 29.2 & 65.4 & 0.70 & 1.01 & 0.140 & 59.7 & 38.0 & 50.3 & 29.1 & 0.81 & 0.23 \\
        CLIP-RN101-LM & 36.2 & 1.320 & 620 & 31.7 & 69.1 & 0.72 & 1.11 & 0.156 & 62.4 & 39.3 & 54.2 & 32.9 & 0.78 & 0.24 \\
        CLIP-ViT-L/14-336px-LM & 32.6 & 1.358 & 923 & 29.8 & 67.1 & 0.70 & 1.01 & 0.138 & 59.0 & 37.6 & 49.1 & 28.4 & 0.57 & 0.87 \\
        GPT-2 & 34.1 & 1.379 & 784 & 29.8 & 67.0 & 0.72 & 1.03 & 0.148 & 60.6 & 38.3 & 50.3 & 28.9 & 0.93 & 0.22 \\        
        \bottomrule
    \end{tabular}
}
\caption{Detailed ablations investigating the impact of various language models on UniAD-based models.}
\label{tab:sup-abl-llm-uniad}
\end{table*}

\section{Long-tail Generalization for 3D Object Detection}
\label{sec:sup-gen}

Tab.~\ref{tab:sup-long-tail-3d} presents the generalization ability of each VLP component on long-tail cases for 3D Object Detection. The results highlight the efficacy of each component in mitigating the long-tail detection problem. Particularly, the inclusion of SLP leads to a noticeable improvement over the baseline, and the combined utilization of both SLP and ALP further enhances the generalization ability. The long-tail classes, including construction vehicles, buses, motorcycles, bicycles, and trailers, constitute approximately 6\% of the nuScenes dataset.

\section{Visualization}
\label{sec:vis}

\begin{figure*}[bt!]
  \centering\includegraphics[width=1.0\linewidth]{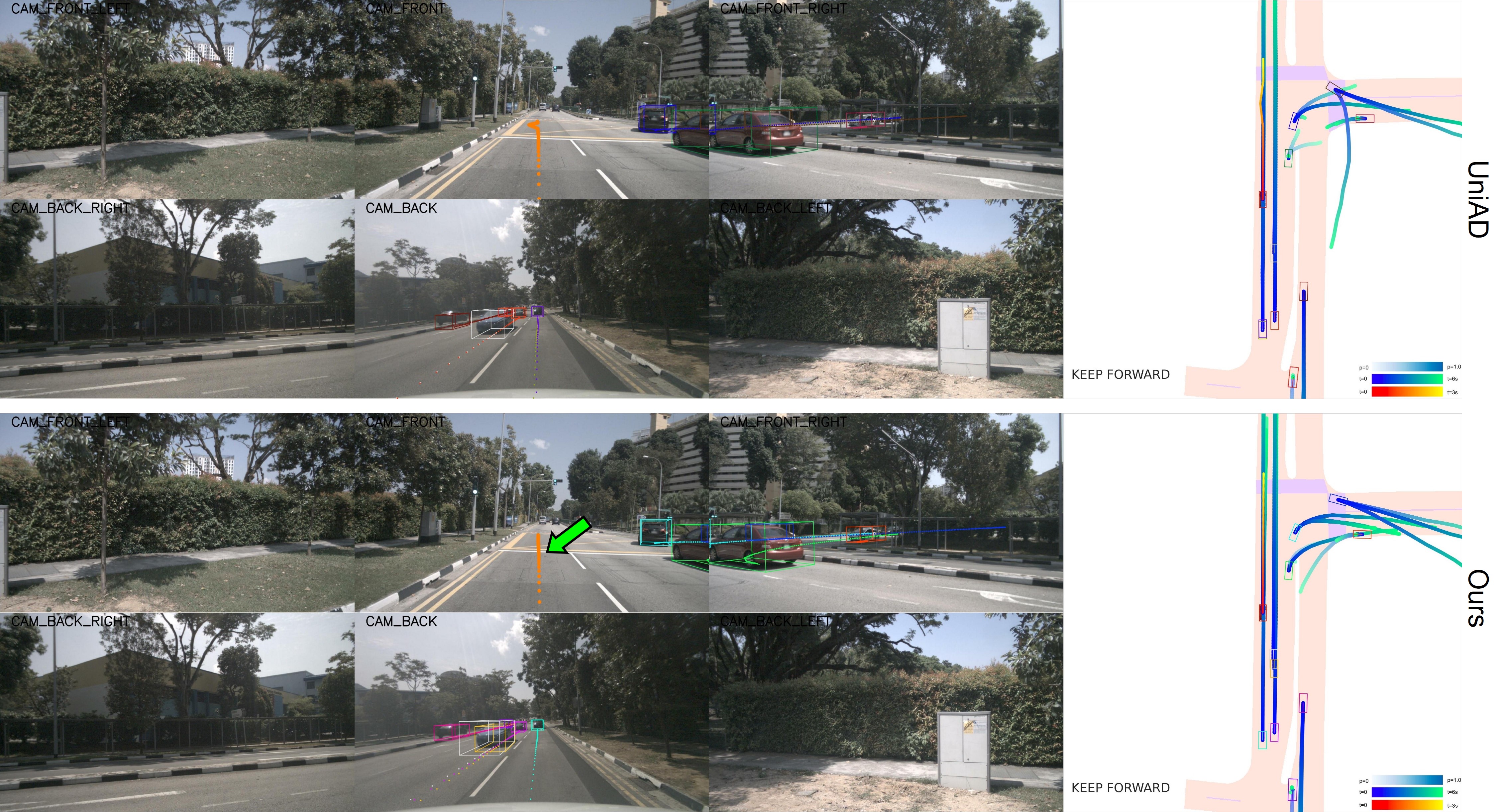}
   \caption{\textbf{Qualitative comparison between UniAD and Ours.} Green arrow is used to highlight areas where our VLP outperforms the baseline. The results indicate that our VLP enables the self-driving car to navigate more efficiently and safely.}
   \label{fig:sup-vis-1}
\end{figure*}

\begin{figure*}[bt!]
  \centering\includegraphics[width=1.0\linewidth]{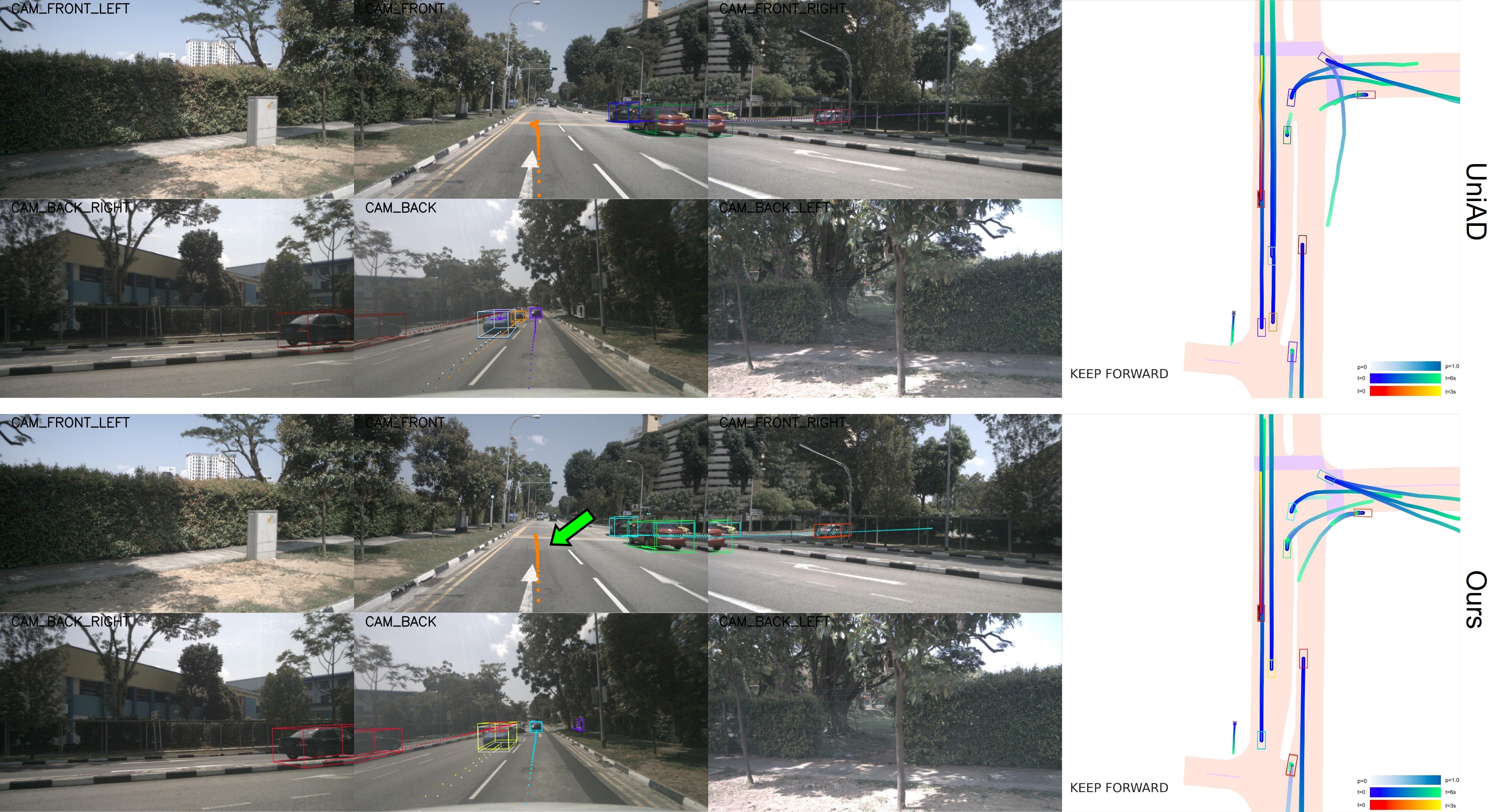}
   \caption{\textbf{Qualitative comparison between UniAD and Ours.} Green arrow highlights areas where our VLP outperforms the baseline.}
   \label{fig:sup-vis-2}
\end{figure*}

\begin{figure*}[bt!]
  \centering\includegraphics[width=1.0\linewidth]{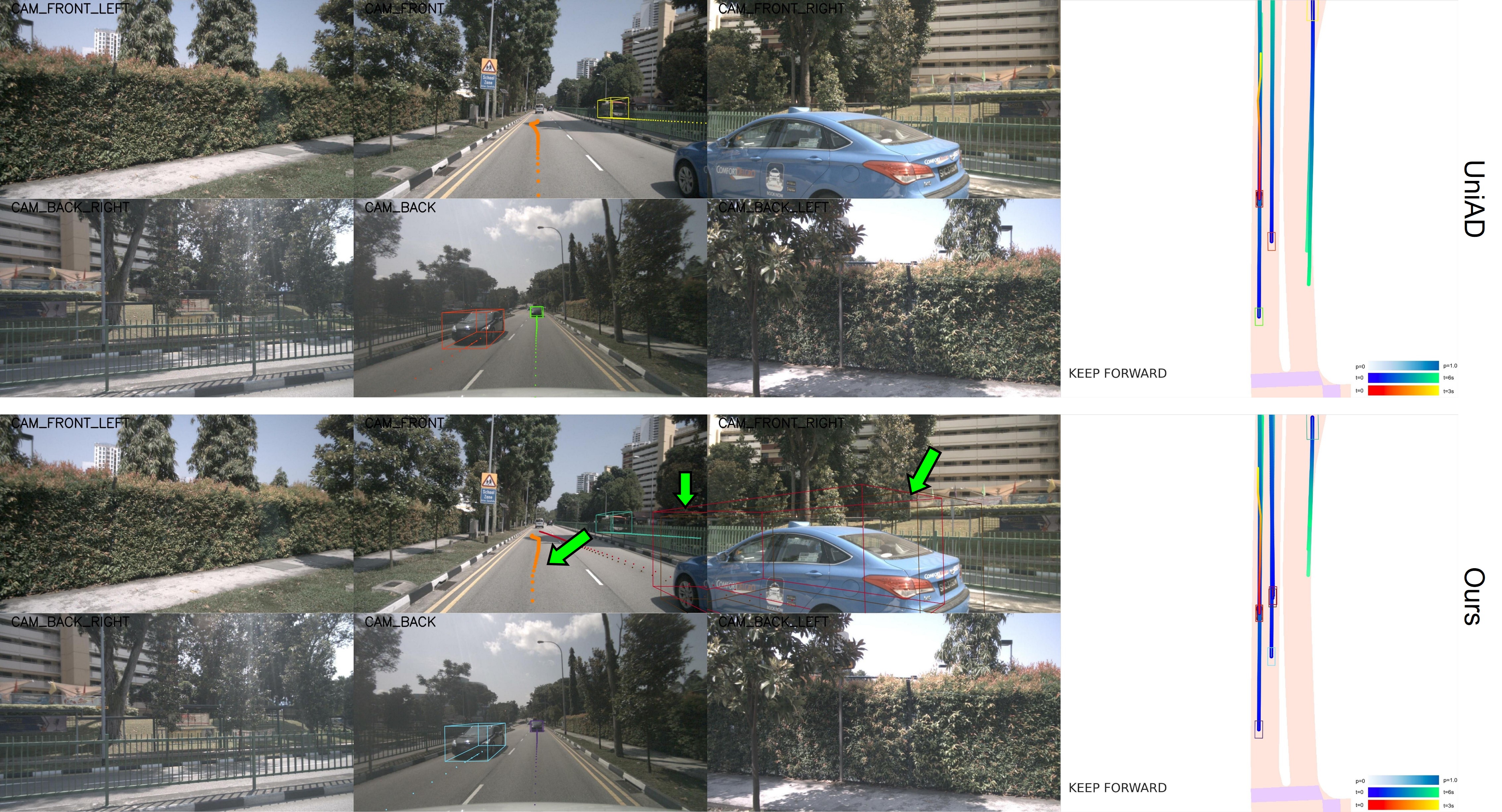}
   \caption{\textbf{Qualitative comparison between UniAD and Ours.} Green arrow highlights areas where our VLP outperforms the baseline.}
   \label{fig:sup-vis-3}
\end{figure*}

\begin{figure*}[bt!]
  \centering\includegraphics[width=1.0\linewidth]{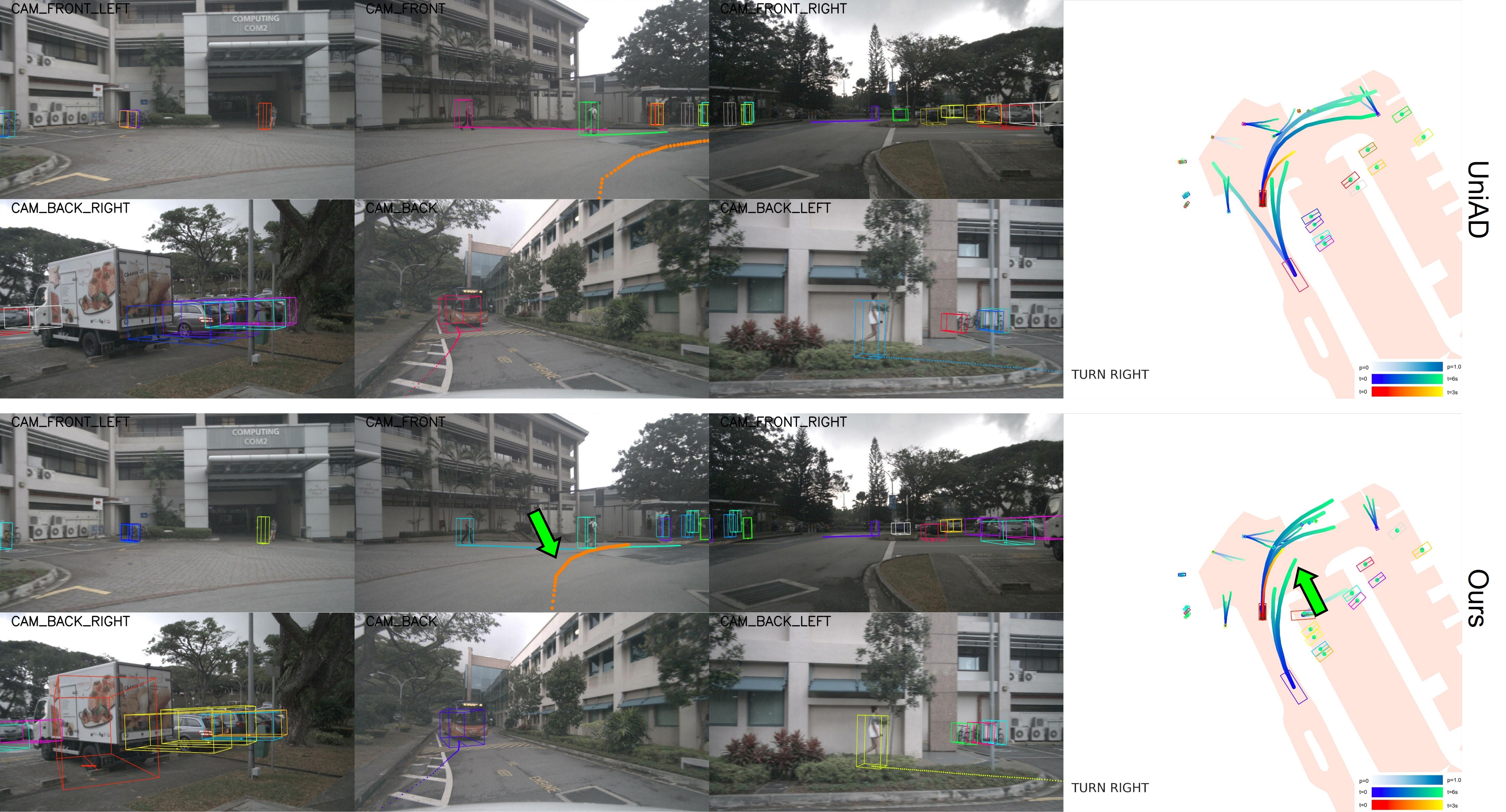}
   \caption{\textbf{Qualitative comparison between UniAD and Ours.} Green arrow highlights areas where our VLP outperforms the baseline.}
   \label{fig:sup-vis-4}
\end{figure*}

\begin{figure*}[bt!]
  \centering\includegraphics[width=1.0\linewidth]{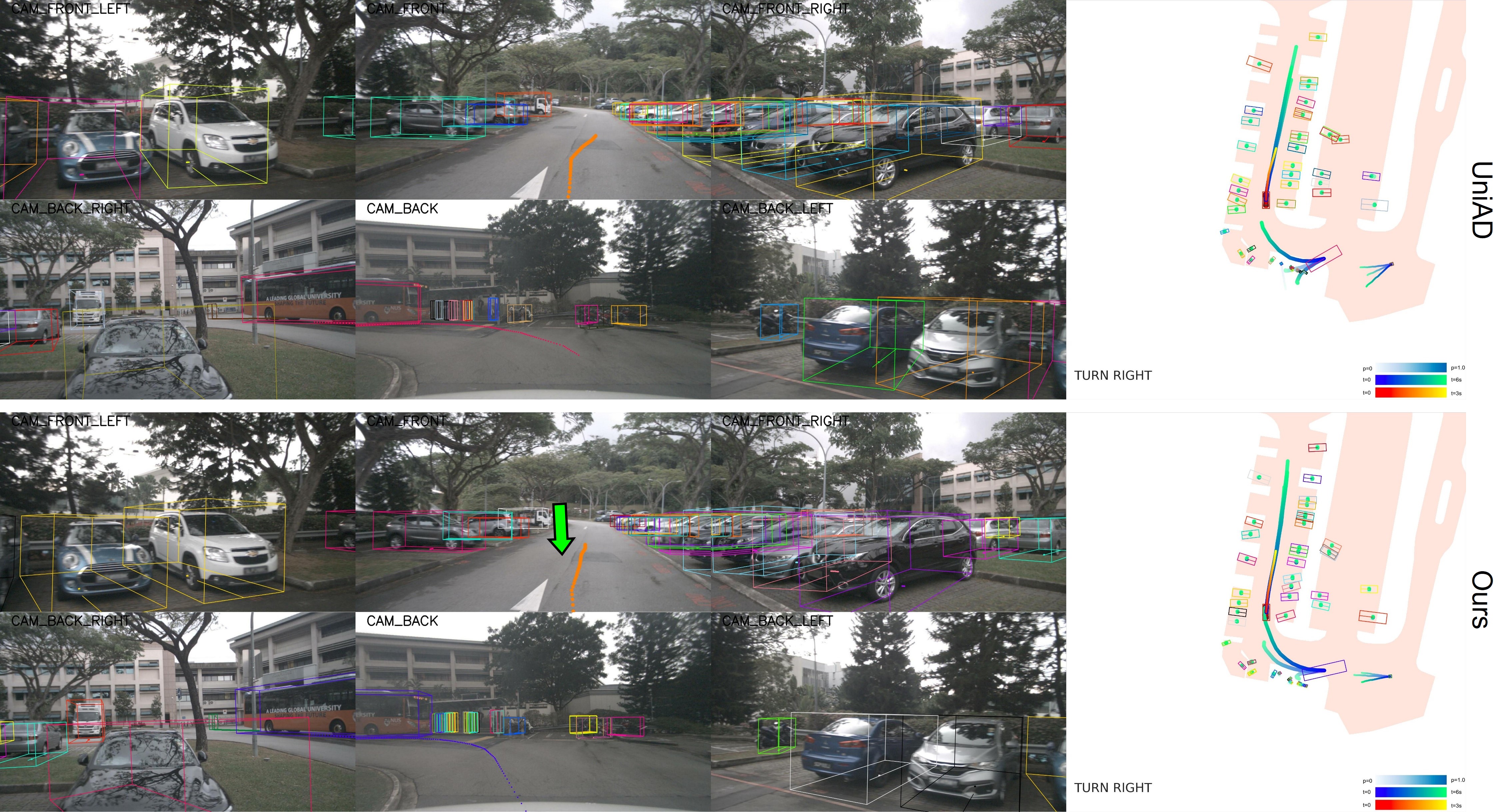}
   \caption{\textbf{Qualitative comparison between UniAD and Ours.} Green arrow highlights areas where our VLP outperforms the baseline.}
   \label{fig:sup-vis-5}
\end{figure*}

\begin{figure*}[bt!]
  \centering\includegraphics[width=1.0\linewidth]{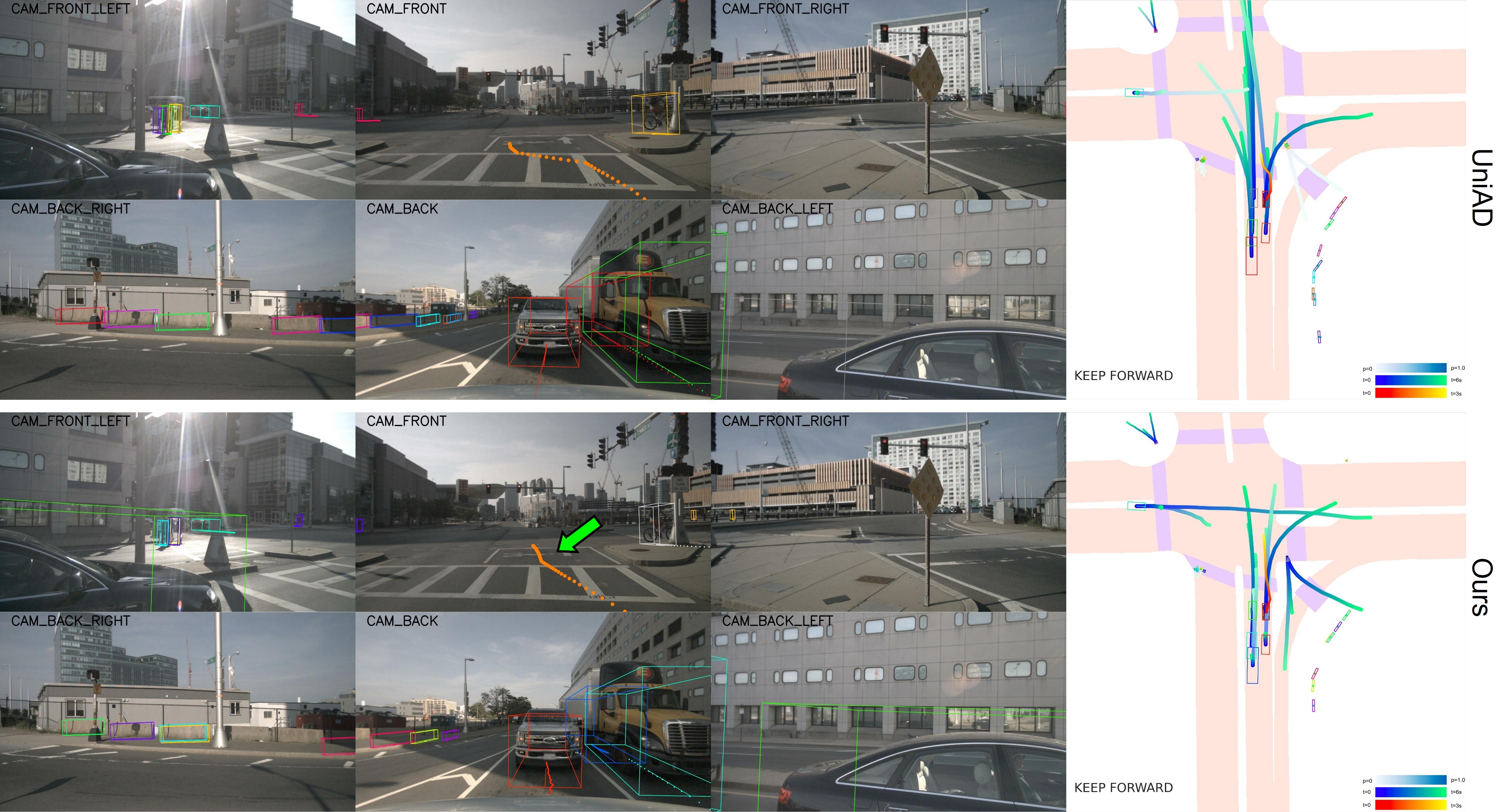}
   \caption{\textbf{Qualitative comparison between UniAD and Ours.} Green arrow highlights areas where our VLP outperforms the baseline.}
   \label{fig:sup-vis-6}
\end{figure*}

\begin{figure*}[bt!]
  \centering\includegraphics[width=1.0\linewidth]{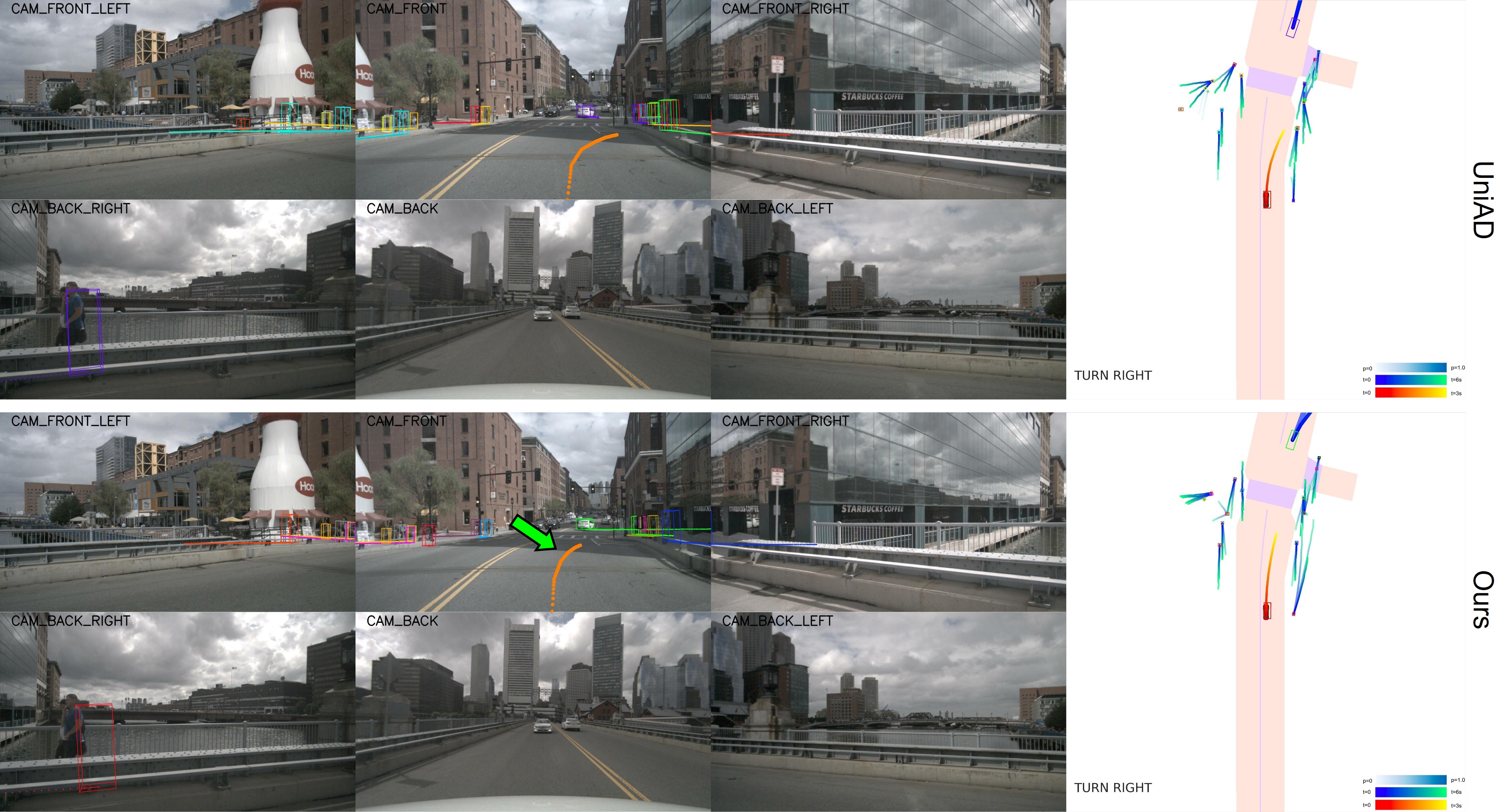}
   \caption{\textbf{Qualitative comparison between UniAD and Ours.} Green arrow highlights areas where our VLP outperforms the baseline.}
   \label{fig:sup-vis-7}
\end{figure*}

We present several qualitative comparisons with the baseline in Fig. \ref{fig:sup-vis-1}-\ref{fig:sup-vis-7}, using green arrows to highlight areas where our model outperforms the baseline. The visual comparison illustrates that our Vision Language Planning (VLP) framework help navigate the self-driving car in a more efficient and safer way.

\section{Why human-like?}
Humans interpret the visual scene as contextual cognitive semantics instead of plain digits, enabling them to navigate in unseen environments and recognize rare objects. We imbue the visual system with a more human-like and intuitively expected feature space via inspiring the reasoning and decision-making processes with designed prompts and pretrained LMs. During inference, our system retains the same enriched feature space and robust capabilities, which is evidenced by the marked improvements in the new-city generalization and long-tail tracking/detection. 

